# Streamlined Photoacoustic Image Processing with Foundation Models: A Training-Free Solution

**Handi Deng,**[1,2,3,†] **Yucheng Zhou,**[4,†] **Jiaxuan Xiang,**[5] **Liujie Gu,**[1,2,3] **Yan Luo,**[1] **Hai Feng,**[6] **Mingyuan Liu,**[6,*] **Cheng Ma**[1,2,3*]

[1]*Beijing National Research Center for Information Science and Technology, Department of Electronic Engineering, Tsinghua University, 30 Shuangqing Road, HaiDian District, Beijing 100084, China*
[2]*Institute for Precision Healthcare, Tsinghua University, 77 Shuangqing Road, HaiDian District, Beijing 100084, China*
[3]*Institute for Intelligent Healthcare, Tsinghua University,77 Shuangqing Road, HaiDian District, Beijing 100084, China*
[4]*School of Biological Science and Medical Engineering, Beihang University, 37 XueYuan Road, HaiDian District,Beijing,100191,China*
[5] *TsingPAI Technology Co., Ltd., 27 Jiancaicheng Middle Road, HaiDian District, Beijing,100096, China*
[6]*Department of Vascular Surgery, Beijing Friendship Hospital, Capital Medical University, 95 Yongan Road, HaiDian District, Beijing 100050, China*
*\* dr.mingyuanliu@pku.edu.cn*
*\* cheng_ma@tsinghua.edu.cn*
*† These authors contribute equally.*

**Abstract**: Foundation models have rapidly evolved and have achieved significant accomplishments in computer vision tasks. Specifically, the prompt mechanism conveniently allows users to integrate image prior information into the model, making it possible to apply models without any training. Therefore, we propose a method based on foundation models and zero training to solve the tasks of photoacoustic (PA) image segmentation. We employed the segment anything model (SAM) by setting simple prompts and integrating the model's outputs with prior knowledge of the imaged objects to accomplish various tasks, including: (1) removing the skin signal in three-dimensional PA image rendering; (2) dual speed-of-sound reconstruction, and (3) segmentation of finger blood vessels. Through these demonstrations, we have concluded that deep learning can be directly applied in PA imaging without the requirement for network design and training. This potentially allows for a hands-on, convenient approach to achieving efficient and accurate segmentation of PA images. This letter serves as a comprehensive tutorial, facilitating the mastery of the technique through the provision of code and sample datasets.

Github: Adi-Deng/photoacoustic-SAM (github.com)

## 1. Introduction

Foundation models (FMs) have flourished with their parameters increasing to hundreds of billions or even trillions[1,2]. Due to the substantial development of parameters, data, and computational power, FMs have demonstrated extraordinary capabilities in numerous tasks. Notably, the natural language processing FMs represented by ChatGPT have shown astonishing abilities in language understanding, generation, inference, and various code-related tasks[3]. They have been widely applied in various fields such as office software, chatbots, translation, text generation, and even assisting medical diagnosis[4]. The development of FMs in computer vision follows closely[5]. Vision Transformer (ViT) model applies the transformer structure into image recognition tasks, significantly increasing the parameter size of vision models[6]. Contrastive Language-Image Pretraining (CLIP) trains vision models using text as prompt, achieving zero-shot classification[7]. Beyond text prompt, some research efforts are also dedicated to using visual prompt. Recently, META's Segment Anything Model (SAM) has demonstrated robust generalization capabilities in segmenting natural images[8], by effectively processing both images and visual prompts (such as boxes, dots, or masks). Moreover, this FM does not require high computational power and can be deployed on ordinary consumer-grade GPUs, offering good prospects for practical applications. Overall, visual FMs have three characteristics: (1) strong generalization ability, allowing a single model to complete tasks in various scenarios; (2) the ability to introduce image prior information through "prompts," simplifying or even avoiding the cumbersome training process. (3) the models do not require high computational power and can be conveniently integrated into imaging hardware.

Prominently, the introduction of prompts enables the incorporation of image prior information into deep learning (DL) models, marking a departure from the traditional approach of designing, improving, and training networks for specific tasks. This paradigm shift reduces the technical hurdles to accessing DL models significantly. In this new paradigm, formulating precise and efficient prompts becomes a research-worthy challenge, recognized as prompt engineering[9].

Image segmentation is a common image processing task in photoacoustic imaging (PAI)[10], with applications spanning vascular segmentation[11-15], tissue boundary delineation[16-18], outer contour segmentation of imaged objects[19-21], and the identification of surgical instruments[22]. In the above-mentioned applications, commonly used segmentation methods include manual segmentation, graphics methods, and DL methods. Among them, DL has emerged as the mainstream approach, yet, it still exhibits certain limitations, with two major ones identified below:

(1) The implementation of traditional DL models involves designing networks, constructing datasets, training, and fine-tuning the network, which require a considerable amount of time and effort[23]. Some researchers may resort to manual image segmentation due to a lack of necessary resources for building datasets.

(2) DL networks previously developed for PA image processing are highly specialized and lack the necessary generalizability to be widely implemented across various imaging scenarios.

In response to these challenges, we report a method, abbreviated as SAMPA (SAM-assisted PA image processing), for zero-training PA image processing based on the SAM FM. Prior knowledge of the imaged object can be conveniently integrated into the modal through prompts and utilized in downstream processing of the segmentation results. The outstanding generalizability of SAMPA was validated through three demonstrations, wherein the imaging systems and objects were deliberately selected to be highly diverse:
(1) Demonstration 1: Removing the skin signal in three-dimensional (3D) PA image rendering. In 3D human hand imaging, SAMPA was used to delineate human tissue boundary and remove signals from the skin, thereby effectively exposing deeper vascular features.
(2) Demonstration 2: Dual speed-of-sound (SoS) reconstruction. In two-dimensional (2D) mouse imaging, SAMPA identified the boundary between the animal and the coupling medium to facilitate dual SoS reconstruction.
(3) Demonstration 3: Human finger's blood vessel segmentation. In the segmentation task, SAMPA robustly identified major blood vessels by refining SAM's output through the incorporation of prior information into a simple algorithm.

In all three tasks, we did not prepare datasets or perform any model training. Instead, we directly deployed the SAM model and combined it with prior information to achieve good results, demonstrating the exceptional simplicity and generalizability of SAMPA. This paves the way for the application of DL in PA image processing and establishes a new standard against which different DL methods can be compared. The objective of this paper is to offer a tutorial, with publicly available code and exemplifying data files, for swiftly implementing SAMPA.

## 2. Materials and Methods

In this section, we introduce the basic workflow of SAMPA, aiming to give readers a fundamental understanding of the approach. In the current paper, we mainly elaborate on emphasizing the simplicity of the method. Detailed module and tool definitions as well as implementation details are provided in the GitHub repository (Adi-Deng/photoacoustic-SAM, github.com). Even readers with no background in DL can quickly replicate and expand upon this work.

### 2.1. *Algorithm Workflow*

The method workflow is illustrated in Figure 1 and consists of two main steps: FM image segmentation custom processing by the boundary information.

In the first step, the primary task is to segment the image using the SAM model based on the image and prompt information. The prompt information is conveyed to the FM through marked points on the image and the category of the area where the marked points are located. The SAM model outputs binary (or multi-value) boundary information. Notably, the SAM model has high compatibility with images of different sizes, and there is generally no need to resize the images, as is often required by many DL methods during a pre-processing step. In Demonstrations 1 and 2, image segmentation involves the division between the imaging object and the coupling medium, thus SAMPA outputs a binary image. In Demonstration 3, the output for blood vessel segmentation is a multi-valued mask which highlights different regions with distinct colors corresponding to their respective categories.

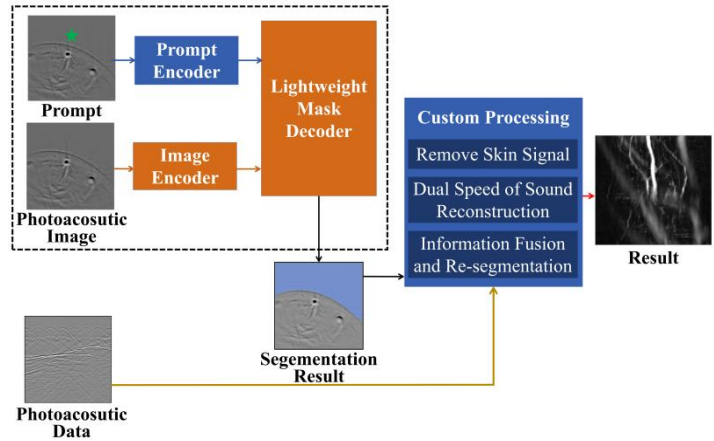

Figure 1: Schematic diagram of SAMPA. The SAM model is outlined with a black dashed box. This method utilizes the segmentation results from the SAM model to achieve more accurate reconstruction or processing of PA data, thereby obtaining better imaging results.

The second step involves "custom processing" which aims to enhance image quality or improve segmentation accuracy based on SAM's output. In this step, specific prior information relevant to the imaging task is incorporated, and various processing methods can be customized. In Demonstration 1, the veins on the back of the human hand are predominantly located close to the surface. Therefore, we generated a mask to filter out the skin's signal and isolate deeper image features, which are primarily contaminated by artifacts. The mask was delineated from the upper boundary of the imaged object to a depth of 1 cm below this boundary. Within the masked region, the value was set to 1, while outside this region, it was set to 0. This process effectively suppressed skin signals and reflection artifacts. In Demonstration 2, the reconstruction process requires determining the time-of-flight (ToF) of the PA signal. For dual-SoS reconstruction, an analytical expression of the body's outline was necessary. Based on SAM's segmentation result, it is straightforward to determine the best elliptical fit to the body's profile. In Demonstration 3, SAM was initially used to identify blood vessels. However, the segmentation results also included non-vessel features. In the second step, we refined the segmentation results by developing a program to calculate the area of each segmented region and the average signal intensity within it.

Depending on the type of the output mask (binary or multi-valued), two sets of code were developed. Demonstrations 1 and 2 shared one set of code, while Demonstration 3 used the other set.

## 2.2. Algorithm Deployment

The computer used in this study consists of a 13th Gen Intel(R) Core(TM) i7-13700K CPU, a GIGABYTE GTX 1660 Super 6G graphics card, and 32G of Kingston DDR4 2666 RAM. Running the lightweight version of the SAM model on Windows takes approximately 0.07 seconds to perform binary segmentation on an image with dimensions of 500×500 pixels. All image segmentation experiments were conducted using the aforementioned hardware and software. The deployment of the model, along with a basic explanation of each component, is detailed in the "How to start" section of our GitHub repository. The process for using the method is described in the "readme" section, with some code and procedural explanations referencing the official SAM repository. The used checkpoint is "sam_vit_l".

## 2.3. Imaging Experiments

We collected PA images of the hand and forearm of a healthy volunteer by a clinical PAI platform (CPIIP, TsingPAI Co., Ltd). The ultrasound probe (256 elements, 5 MHz center frequency, 60% receive bandwidth) had a 180° angular coverage providing 2D cross-sectional images. An optical parametric oscillator (OPO) provides excitation pulses at 850 nm. The scanning step was 0.1 mm. A 3D image was reconstructed by splicing 2D cross-sectional images. A healthy volunteer (20-year-old male, Asian, moderate skin tone) participated in the test.

Small animal imaging was performed by a custom-made ring array PA computed tomography (PACT) system with 256 transducer elements, 5 MHz center frequency and 70% receive bandwidth. A rotation of the array by 0.7 degrees resulted in an equivalent acquisition of 512 channels. The excitation wavelength was 850nm. An 8-week-old female NU/NU mouse (SPF) was used as the imaging subject.

Human finger imaging was performed using the aforementioned ring-array PACT system without rotating the array. Additionally, only half of the ring array data were utilized for image reconstruction, resulting in a 128-channel half-ring acquisition. We intentionally employed this limited-angle acquisition to induce limited-view artifacts, thereby increasing the complexity of the segmentation task. The excitation wavelength was 800 nm. A healthy volunteer (23-year-old male, Asian, moderate skin tone) participated in the imaging experiment.

In all experiment, we maintained a laser repetition rate of 10 Hz and ensured that the per-pulse energy density remained below 15 mJ/cm$^2$ to comply with the limits set by the American National Standards Institute (ANSI). The images were reconstructed by a standard delay and sum (DAS) algorithm. Both the hand, finger and the mouse were submerged in distilled water to facilitate ultrasound coupling.

The animal study was reviewed and approved by National Institutes of Health Guidelines on the Care and Use of Laboratory Animal of Beijing Vital River Laboratory Animal Technology Co., Ltd. The human experiments have been approved by the Ethics Committee of Tsinghua University (Project No: 20220121).

## 3. Results

In this section, we will first elaborate on setting the prompts. Then, we will demonstrate the effectiveness of SAMPA.

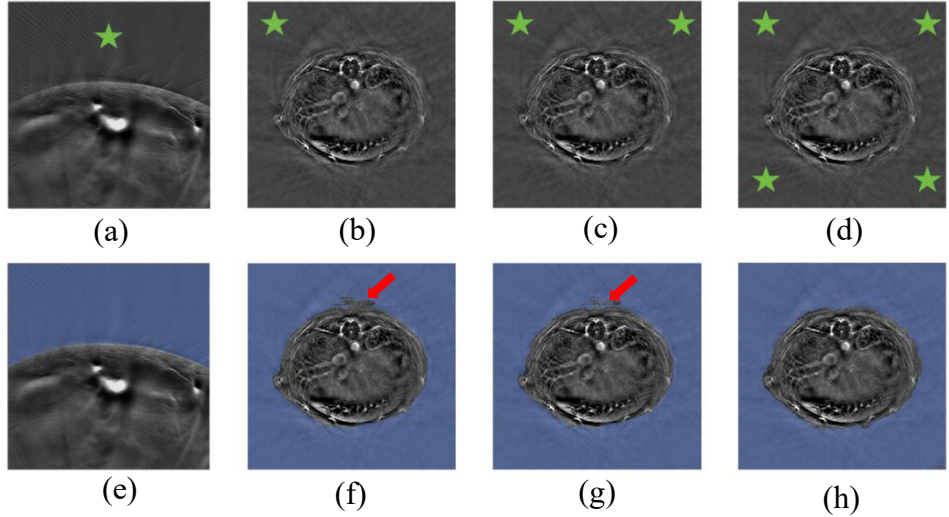

Figure 2: Results of human hand and mouse body segmentation under different prompts (represented by green stars). (a) and (e) show the original PA image of the hand and the segmentation result, respectively. In this demonstration, a single prompt yielded good results. (b) and (f) display the original mouse image and the segmentation result, respectively. A single prompt was used. (c), (f) and (d), (h) illustrate updated results, when two and four prompts were used, respectively. The red arrows point to areas with incorrect segmentation.

In Demonstration 1, given that the coupling medium is located above the hand, we marked the position of the coupling medium for all images acquired at different scanning positions. The position of the hand relative to the surrounding medium during the scanning procedure was consistent. Therefore, we marked the same prompt point for each scanning position.

In the mouse imaging experiment, to challenge the segmentation task, we deliberately selected a 2D layer where signals from the internal organs were significantly stronger than those from the skin, making it difficult to distinguish the skin boundary.

Figure 2 shows the segmentation results of the human hand and the mouse body under different prompts. It can be observed that satisfactory segmentation can be achieved with only a few prompt points, without any fine-tuning. In the hand segmentation task shown in Figures 2(a) and (e), an accurate segmentation result was obtained with only one prompt point, indicating good generalization capability. The segmentation quality for the mouse body is consistently good. As shown in Figures 2(f) and (g), when the number of prompts was two or fewer, a small region at the top portion of the body was incorrectly segmented, as indicated by the red arrows. With four prompts, the aforementioned segmentation error was corrected, as shown in Figure 2(h). This has led to the conclusion that moderately increasing the number of prompts can effectively improve segmentation accuracy.

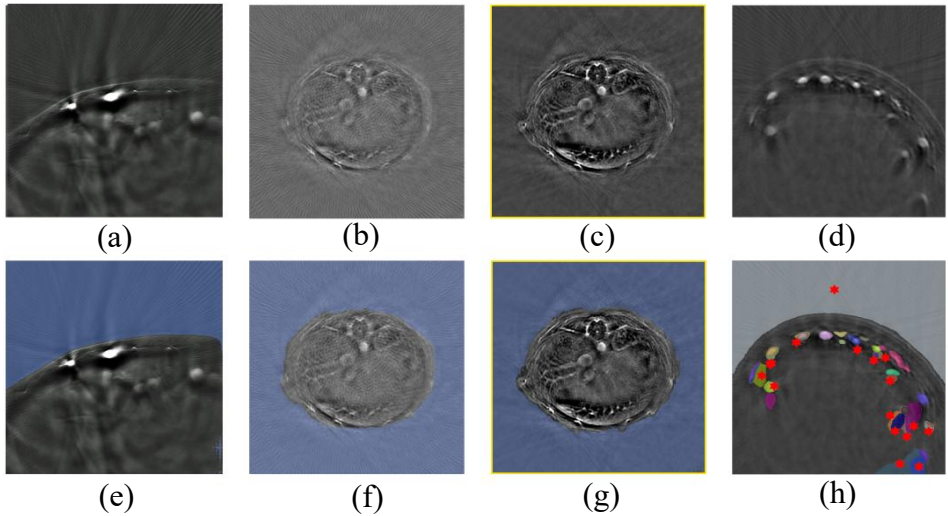

Figure 3: Image segmentation results in relatively complex tasks. (a) and (e) show the segmentation results under limited-view artifacts. (b) and (f) are the segmentation results under sparse-sampling. Compared with the fully-sampled images shown in (c) and (g) (highlighted with yellow boxes), it is evident that (f) is well segmented despite being under-sampled. (d) and (h) display the multi-value vessel segmentation results, revealing some apparent misclassifications, labelled by red hexagrams, that necessitate further refinement of the segmentation results.

The dimensions of the hand image are $500 \times 500$ pixels, and the runtime with 1 prompt point is 0.069s. The dimensions of the mouse image are $500 \times 500$ pixels, and the runtimes with 1, 2, 4 prompt points are 0.072s, 0.071s, 0.070s, respectively. The overall runtime of the model does not change significantly with the variation of prompts. The above findings indicate that SAM can perform segmentation of traditional PA images within 0.1s. This capability renders it suitable for deployment within conventional imaging apparatus or seamless integration into PA image processing software.

We also tested the segmentation performance in relatively complex scenarios of Demonstrations 1 and 2, including: (1) Human arm images with limited-view artifacts (Figures 3(a) and (e)); (2) Mouse cross-sectional images reconstructed with under-sampled data, simulating a cost-effective ring-array with only 128 channels. To mitigate streak artifacts, a simple method was employed to expand the data into 256 channels before reconstruction, as illustrated in Figures 3(b) and (f). Specifically, the first channel of the original data was duplicated into the first and second channels of the new data. Subsequently, the second channel was duplicated into the third and fourth channels, and so forth. Figures 3(c) and (g) depict the image reconstructed from fully-sampled data (512 channels) and its segmentation result.

In Figure 3(a), strong artifacts were produced due to the limited acceptance angle of the transducer. However, these artifacts had minimal influence on the accuracy of the segmentation, as evidenced by Figure 3(e). Similarly, Figure 3(b) exhibits poor image quality due to under-sampling, resulting in a vague boundary of the animal. Nevertheless, in Figure 3(f), the segmentation result was comparable to that of the fully-sampled image, as demonstrated in Figure 3(g). These findings suggest that SAM can robustly achieve accurate segmentation despite limited-view and under-sampling conditions.

Figure 3(d) displays the PA image of the human finger, while the blood vessel segmentation result is shown in Figure 3(h). All vessels were successfully identified. However, several image features that are clearly artifacts were incorrectly recognized as blood vessels, as indicated by the red hexagrams, necessitating the additional step of "custom processing."

Figure 4 displays the final reconstruction results of Demonstrations 1, 2 and 3. Figures 4(a) and (e) show the max intensity projection (MIP) images of the 3D blood vessel reconstruction of the human hand, before and after the removal of the skin signals. It is evident that the removal of skin signals, assisted by the aforementioned segmentation, better reveals deeper blood vessels, as pinpointed by the white arrows. Figures 4(b) and (f) show the single- and dual-SoS reconstructions of the mouse trunk, respectively. Figures 4(c) and (g) display the zoomed-in images of the corresponding areas in (b) and (f). The places where image quality has effectively improved are indicated by white circles. While deep features remain invariant, superficial features become more in-focus after dual-SoS reconstruction based on the segmentation result of SAM. Figures 4(d) and (h) depict the cross-sectional PA image of a human finger and the result of blood vessel segmentation. As shown in Figure 4(h), refined segmentation was obtained using a customized program based on the original segmentation result obtained from SAM (Figure 3(h)). It is evident that by integrating simple prior knowledge of the PA image, the segmentation became more accurate.

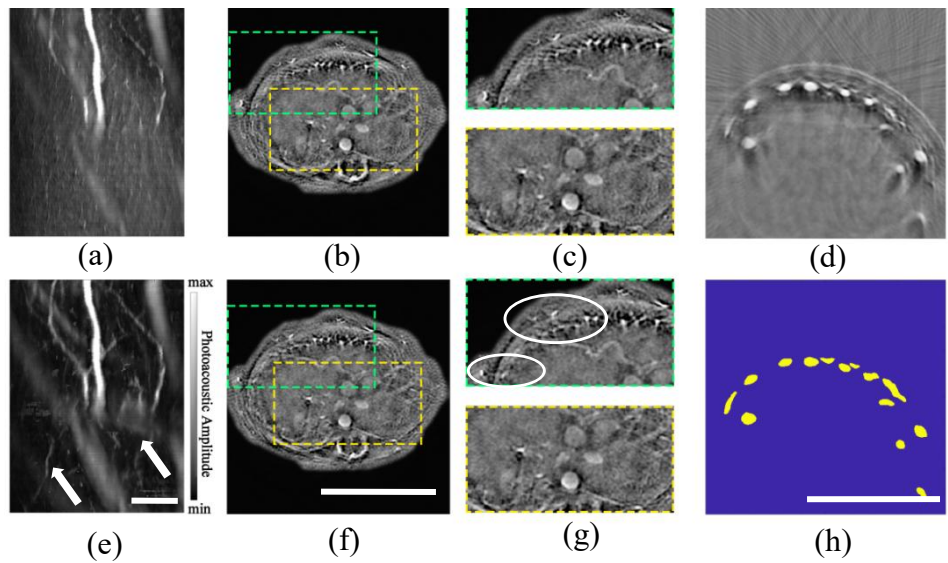

Figure 4: Imaging results of Demonstrations 1, 2 and 3. (a) and (e) show 3D imaging results (maximum intensity projection) before and after the removal of surface signals. Write arrows label the vessels that are better-exposed. (b) and (f) display the results of single- and dual- speed of sound (SoS) reconstruction, respectively. (c) and (g) are magnified images of (b) and (f), showing that both external features (outlined by green dashed boxes) and internal features (outlined by yellow dashed boxes) can be well-reconstructed in the dual-SoS images. White circles in (g) indicate areas with improved image quality. (d) and (h) are photoacoustic images of finger vessels and their segmentation results. Scale bars: 10 mm.

## 4. Discussion and Conclusions

This paper reports a method that applies FMs for segmenting PA images and performing reconstruction, effectively achieving good results across multiple tasks. By implementing SAMPA in a convenient, training-free manner, we demonstrated the usefulness of the method across three imaging scenarios. In human hand imaging, the segmentation of the skin signal and its subsequent removal effectively revealed internal blood vessels. In animal cross-sectional imaging, auto-segmentation of the body profile facilitates dual-SoS reconstruction, thereby enhancing image quality. In human finger imaging, blood vessel segmentation was achieved, potentially aiding in medical diagnosis.

This paper highlights several advantages of visual FMs: (1) Training-free: The method of integrating prior information through prompts allows for the direct application of DL models without pre-training. (2) Robustness: The model achieves good segmentation results even in the presence of artifacts, greatly enhancing its applicability in complex real-world scenarios. (3) Efficiency: Compared to language FMs, vision FMs have lower computational requirements, greatly facilitating deployment. In traditional DL practices, designing networks and preparing datasets are time-consuming tasks, and the quality of the training set critically determines the model's performance. In contrast, FMs removes these technical barriers, enabling researchers to implement DL models quickly and conveniently. Some FMs, such as SAM, provide online demos, making model verification and deployment readily accessible. Freed from the tasks of dataset preparation and network design, the only remaining task for the user is to design appropriate prompts to apply FMs effectively. Such favorable properties contribute to the uniqueness of FMs in 3D image processing, which is encountered routinely in medical applications. Since the workload for network design and dataset preparation significantly increases for 3D image processing, a training-free, plug-and-play model can expect to receive widespread adoption [24]. We envision that in the future, directly implementing FMs, or performing simple fine-tuning on the basis of FMs, is a promising solution to applying DL in PAI.

In summary, applying FMs has proven to be a promising solution for implementing DL in PAI. It is worth mentioning that SAM's online trial offers convenient hands-on practice for quickly mastering the program. To facilitate readers in replicating our results and validating our findings, we have uploaded all of our codes and provided detailed documentation on https://github.com/Adi-Deng/photoacoustic-SAM.

## Acknowledgments

We would like to acknowledge financial support from Strategic Project of Precision Surgery, Tsinghua University; Initiative Scientific Research Program, Institute for Intelligent Healthcare, Tsinghua University; Tsinghua-Foshan Institute of Advanced Manufacturing; National Natural Science Foundation of China (61735016); Beijing Nova Program (20230484308); Young Elite Scientists Sponsorship Program by CAST (2023QNRC001);Youth Elite Program of Beijing Friendship Hospital (YYQCJH2022-9); Science and Technology Program of Beijing Tongzhou District (KJ2023CX012).

We thank Xiaojun Wang, Yuwen Chen, Wubing Fu, Naiyue Zhang, Wenjie Guo, Jianpan Gao at TsingPAI Technology Co., LTD. for helpful discussions.

## Conflicts of Interest

C.M. had a financial interest in TsingPAI Technology Co., LTD., which provided the clinical imaging system (CPIIP) used in this work.

## References

1. Bommasani, Rishi, et al. "On the opportunities and risks of foundation models." arXiv preprint arXiv:2108.07258 (2021).
2. Zhou, Ce, et al. "A comprehensive survey on pretrained foundation models: A history from bert to chatgpt." arXiv preprint arXiv:2302.09419 (2023).
3. Wu, Tianyu, et al. "A brief overview of ChatGPT: The history, status quo and potential future development." IEEE/CAA Journal of Automatica Sinica 10.5 (2023): 1122-1136.
4. Biswas, Som S. "Role of chat gpt in public health." Annals of biomedical engineering 51.5 (2023): 868-869.
5. Awais, Muhammad, et al. "Foundational models defining a new era in vision: A survey and outlook." arXiv preprint arXiv:2307.13721 (2023).
6. Dosovitskiy, Alexey, et al. "An image is worth 16x16 words: Transformers for image recognition at scale." arXiv preprint arXiv:2010.11929 (2020).
7. Radford, Alec, et al. "Learning transferable visual models from natural language supervision." International conference on machine learning. PMLR, 2021.
8. Kirillov, Alexander, et al. "Segment anything." Proceedings of the IEEE/CVF International Conference on Computer Vision. 2023.
9. Wang, Jiaqi, et al. "Review of large vision models and visual prompt engineering." Meta-Radiology (2023): 100047.
10. Le, Thanh Dat, Seong-Young Kwon, and Changho Lee. "Segmentation and quantitative analysis of photoacoustic imaging: a review." Photonics. Vol. 9. No. 3. MDPI, 2022.
11. Tang, Kaiyi, et al. "Advanced image post-processing methods for photoacoustic tomography: A review." Photonics. Vol. 10. No. 7. MDPI, 2023.
12. Yuan, Alan Yilun, et al. "Hybrid deep learning network for vascular segmentation in photoacoustic imaging." Biomedical Optics Express 11.11 (2020): 6445-6457.
13. Vaiyapuri, Thavavel, et al. "Design of metaheuristic optimization-based vascular segmentation techniques for photoacoustic images." Contrast Media & Molecular Imaging 2022 (2022).

14. Ly, Cao Duong, et al. "Full-view in vivo skin and blood vessels profile segmentation in photoacoustic imaging based on deep learning." Photoacoustics 25 (2022): 100310.
15. Gao, Ya, et al. "Deep learning-based photoacoustic imaging of vascular network through thick porous media." IEEE Transactions on Medical Imaging 41.8 (2022): 2191-2204.
16. Sun, Mingjian, et al. "Full three-dimensional segmentation and quantification of tumor vessels for photoacoustic images." Photoacoustics 20 (2020): 100212.
17. Zhang, Jiayao, et al. "Photoacoustic image classification and segmentation of breast cancer: a feasibility study." IEEE Access 7 (2018): 5457-5466.
18. Jnawali, Kamal, et al. "Automatic cancer tissue detection using multispectral photoacoustic imaging." International Journal of Computer Assisted Radiology and Surgery 15 (2020): 309-320.
19. Mandal, Subhamoy, Xosé Luís Deán-Ben, and Daniel Razansky. "Visual quality enhancement in optoacoustic tomography using active contour segmentation priors." IEEE transactions on medical imaging 35.10 (2016): 2209-2217.
20. Li, Lei, et al. "Single-impulse panoramic photoacoustic computed tomography of small-animal whole-body dynamics at high spatiotemporal resolution." Nature biomedical engineering 1.5 (2017): 0071.
21. Huang, Chuqin, et al. "Dual-Scan Photoacoustic Tomography for the Imaging of Vascular Structure on Foot." IEEE Transactions on Ultrasonics, Ferroelectrics, and Frequency Control (2023).
22. Lin, Xiangwei, et al. "Handheld interventional ultrasound/photoacoustic puncture needle navigation based on deep learning segmentation." Biomedical Optics Express 14.11 (2023): 5979-5993.
23. Davoudi, Neda, Xosé Luís Deán-Ben, and Daniel Razansky. "Deep learning optoacoustic tomography with sparse data." Nature Machine Intelligence 1.10 (2019): 453-460.
24. Wang H, Guo S, Ye J, et al. Sam-med3d[J]. arxiv preprint arxiv:2310.15161, 2023.